\documentclass[11pt,a4paper]{article}
\usepackage[fontset=fandol,heading=false,scheme=plain]{ctex}
\usepackage[margin=2.4cm]{geometry}
\usepackage{amsmath,amssymb}
\usepackage{booktabs,graphicx,array,multirow,adjustbox}
\usepackage{pgfplots,pgfplotstable}
\usepgfplotslibrary{groupplots}
\pgfplotsset{compat=1.18}
\usepackage[numbers,sort&compress]{natbib}
\usepackage{xcolor}
\usepackage[colorlinks=true,linkcolor=blue!60!black,citecolor=blue!60!black,urlcolor=blue!60!black]{hyperref}
\usepackage{caption}
\xeCJKsetup{CJKmath=true}
\newcommand{\paperTitle}{How Many Bytes Has a BPE Model Seen? Estimating Byte-Token Exposure in Pretraining from Corpus Statistics and Embedding Norms}
\newcommand{\paperTitleZh}{BPE 模型在预训练里见过多少字节?从语料统计与 embedding 范数两路估计字节 token 的曝光量}
\newcommand{\paperAuthor}{Liwei Zhang \qquad ZhiMing Tong \qquad Mochen Guan \qquad Linzhe Li$^{*}$\\[0.3em]
\normalsize Landi Lab \qquad $^{*}$Corresponding author\\
\normalsize\texttt{z@metiy.xyz} \quad \texttt{mynameistong@163.com} \quad \texttt{gmc@antiy.cn} \quad \texttt{lilinzhe@antiy.cn}}
\newcommand{\paperAuthorZh}{张栗伟(Liwei Zhang) \quad 童志明(ZhiMing Tong)\\[0.2em]
关墨辰(Mochen Guan) \quad 李林哲(Linzhe Li)$^{*}$\\[0.4em]
\normalsize 澜砥实验室(Landi Lab) \qquad $^{*}$通讯作者\\
\normalsize\texttt{z@metiy.xyz} \quad \texttt{mynameistong@163.com} \quad \texttt{gmc@antiy.cn} \quad \texttt{lilinzhe@antiy.cn}}

\title{\paperTitle}
\author{\paperAuthor}
\date{November 2026}

\begin{document}

\maketitle
\begin{abstract}
Byte-level BPE vocabularies keep 256 single-byte tokens as a fallback, and converting such a model into a pure byte-level model (as we did for Qwen3.8-27B in earlier work) starts from the embeddings of exactly these 256 rows. How much training have they actually received? We answer this question for Qwen3.8 with two independent routes and cross-check them. Route one tokenizes 540~MB of public corpora from 40 domains in 33 languages (Wikipedia, FineWeb/FineWeb-2, WildChat, OASST2, GitHub code, FineMath) and multiplies the per-domain shares by assumed pretraining mixtures. Single-byte tokens make up 13--36\% of all tokens, but almost all of that is ASCII (digits split per character, punctuation); the genuine UTF-8 fallback bytes 0x80--0xFF account for only 0.15--0.27\% under mainstream mixtures (0.66\% with a 10\% long-tail share), i.e.\ about $2\times10^{9}$ occurrences per $10^{12}$ pretraining tokens, or $10^{8}$--$10^{9}$ per legal high byte at the 36T-token scale reported for Qwen3. Fallback rates span four orders of magnitude across languages (0.003\% Russian, 0.44\% Chinese, 1.3\% Korean, 45\% Amharic) and are set by how well the vocabulary covers each script. Route two reads the row norms of the embedding and output matrices of four open models (Qwen3.8-27B, Qwen3.5-9B, MiniCPM5-2B, MiniMax-M3.1-Flash), using the never-trained padding rows as a zero-exposure baseline and the frequency-known ordinary tokens as a calibration curve. In all four models exactly the 13 bytes that cannot occur in valid UTF-8 (0xC0, 0xC1, 0xF5--0xFF) are still at their initialization, while every legal high byte sits in the high-frequency band; the equivalent frequency read from the norms agrees with the corpus estimate to within a factor of 4 for Qwen3.8-27B and within 0.05 dex for MiniMax. The norm--frequency curve is monotone only up to a model-dependent fold point, so norm-based inversion is a per-model, order-of-magnitude tool. We release all scripts; no private data were used.

\end{abstract}
\noindent\small This document contains the full paper in English followed by the full paper in Chinese (p.~\pageref{sec:zh}). Both versions describe the same experiments and report the same numbers.
\normalsize
\newcommand{\figHbYlabel}{high-byte tokens, \% of all tokens (log)}
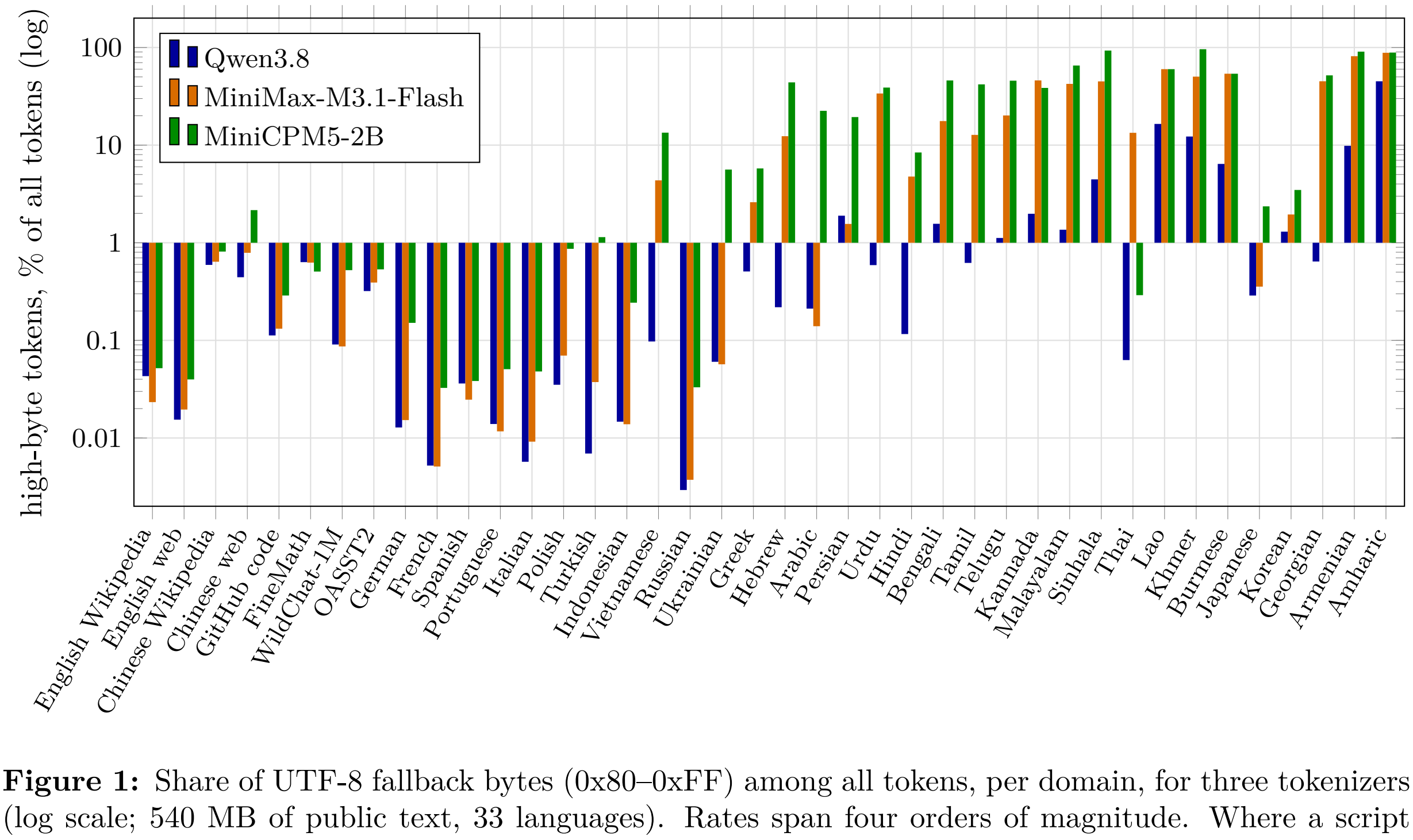
\newcommand{\figHbCaption}{Share of UTF-8 fallback bytes (0x80--0xFF) among all tokens, per domain, for three tokenizers (log scale; 540~MB of public text, 33 languages). Rates span four orders of magnitude. Where a script is absent from a vocabulary, the language becomes effectively byte-level (right side of the plot; Lao and Burmese give identical numbers for MiniMax and MiniCPM5 because both tokenizers fall back on every character).}
\newcommand{\figCurveYlabel}{embedding row norm}
\newcommand{\figCurveLegA}{ordinary tokens (bin median)}
\newcommand{\figCurveLegB}{high bytes 0x80--0xFF}
\newcommand{\figCurveLegC}{ASCII bytes}
\newcommand{\figCurveLegD}{zero-exposure baseline}
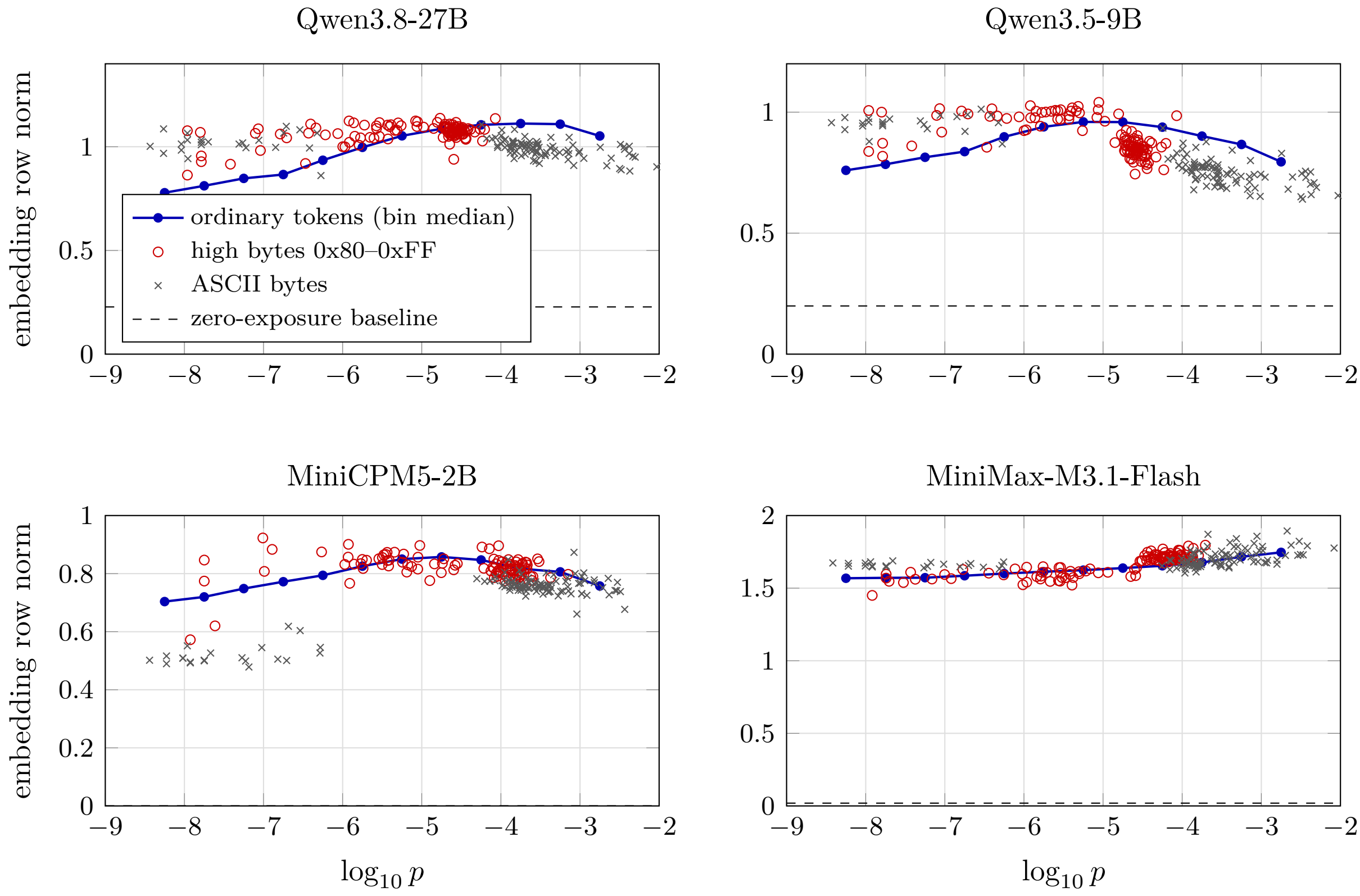
\newcommand{\figCurveCaption}{Embedding row norm versus corpus frequency ($\log_{10}$ of the base-mixture token probability). Blue: median of ordinary multi-byte tokens per 0.5-dex bin; red circles: the 128 high bytes; grey crosses: the 128 ASCII bytes; dashed: the never-trained padding rows. Bytes that never occur in the corpus sample are drawn at the left edge only in the tables (no $x$ value here). The curve is monotone up to a model-dependent fold ($\log_{10}p\approx-4$ for Qwen3.8-27B, $-5$ for Qwen3.5-9B and MiniCPM5-2B, none visible for MiniMax).}

\section{Introduction}
Every GPT-2-style byte-level BPE vocabulary \citep{radford2019gpt2,sennrich2016bpe} contains 256 single-byte tokens. They guarantee that any UTF-8 string can be encoded: a character that has no token of its own is split into its bytes. In our earlier work \citep{zhang2026onebyte} we converted Qwen3.8-27B into a pure byte-level model by discarding all other rows of the embedding and output matrices and keeping exactly these 256 (plus special symbols), and we observed that the conversion did not start ``from scratch'': ASCII content lost almost nothing, and even Chinese recovered within a few hundred million bytes of training. The natural question is \emph{how much training these 256 rows had already received}. The pretraining corpora of production models are not public, so the question cannot be answered by counting; it has to be estimated.

This report makes the estimate in two independent ways and checks whether they agree.
\begin{enumerate}
\item \textbf{Corpus route.} We tokenize 540~MB of public text from 40 domains in 33 languages with the model's own tokenizer, measure how often each of the 256 byte tokens appears, classify the Unicode characters that trigger a fallback, and multiply the per-domain shares by several assumed pretraining mixtures. Where a total pretraining size has been disclosed we convert shares to counts; where it has not, we report counts per $10^{12}$ tokens.
\item \textbf{Weight route.} We read the row norms of the token embedding and output (\texttt{lm\_head}) matrices of four open models. Vocabulary rows that were padded for alignment and never appeared in any batch give a clean zero-exposure baseline; the ordinary tokens, whose corpus frequencies we know from the first route, give a norm--frequency calibration curve; and the 256 byte rows are placed on that curve.
\end{enumerate}

Two distinctions organise the results. The first is between \emph{ASCII single-byte tokens} (0x00--0x7F), which are ordinary tokens for digits, punctuation, spaces and rare letters, and \emph{high bytes} (0x80--0xFF), which appear only when a multi-byte UTF-8 character is split---the genuine ``byte-level'' signal. The second is between the 243 \emph{legal} high bytes and the 13 byte values that cannot occur in valid UTF-8 at all (0xC0, 0xC1, 0xF5--0xFF).\footnote{0xC0 and 0xC1 would be overlong two-byte lead bytes; 0xF5--0xFF would lead sequences beyond U+10FFFF. All 128 continuation bytes 0x80--0xBF, the lead bytes 0xC2--0xDF, 0xE0--0xEF and 0xF0--0xF4 are legal.}

Our main findings:
\begin{itemize}
\item Single-byte tokens are 13--36\% of all tokens in every domain, because Qwen splits digits one per token and most ASCII punctuation is a single-byte token; across mixtures the share is a stable 18.5--21.9\%. This is ASCII, not UTF-8 fallback (Section~\ref{sec:route1}).
\item Genuine fallback (high bytes) is 0.15--0.27\% of tokens under mainstream mixtures and 0.66\% if 10\% of the mixture is long-tail scripts. Per $10^{12}$ pretraining tokens that is $1.5$--$6.6\times10^{9}$ high-byte tokens; at the 36T-token scale disclosed for Qwen3 \citep{yang2025qwen3}, the median legal high byte is seen $\sim7\times10^{8}$ times, with a spread from $<10^{7}$ to $>10^{9}$.
\item Fallback rates differ by four orders of magnitude across languages and are determined by vocabulary coverage: 0.003\% for Russian, 0.44\% for Chinese (rare characters), 1.3\% for Korean (uncovered syllables), 6--16\% for Burmese, Khmer and Lao, 45\% for Amharic. MiniMax-M3.1-Flash and MiniCPM5-2B, with smaller vocabularies, run many more scripts entirely in bytes (Section~\ref{sec:langs}).
\item In all four models, exactly the 13 illegal bytes are still at their initialization norm, and every legal high byte has a norm in the high-frequency band of the calibration curve (Section~\ref{sec:route2}). This is the hardest point of agreement between the two routes: the corpus never produces these bytes, and the weights confirm that the model never saw them.
\item The norm-based equivalent frequency agrees with the corpus estimate to 0.6~dex for Qwen3.8-27B and 0.05~dex for MiniMax, but the norm--frequency curve folds at a model-dependent point (Qwen3.5-9B and MiniCPM5-2B fold earlier), so norm inversion is only an order-of-magnitude, per-model instrument.
\end{itemize}

\section{Related work}
\paragraph{Byte fallback in subword vocabularies.} GPT-2 \citep{radford2019gpt2} introduced the byte-to-unicode mapping that lets BPE \citep{sennrich2016bpe} operate on bytes; Qwen, MiniCPM and MiniMax all inherit it. Pure byte-level models \citep{xue2022byt5,pagnoni2025blt} and byteification of trained models \citep{minixhofer2025bolmo,zhang2026onebyte} make the byte rows the whole vocabulary; our question is what those rows contain before conversion.

\paragraph{Under-trained and glitch tokens.} \citet{rumbelow2023solidgold} found tokens that elicit anomalous behaviour; \citet{land2024magikarp} showed that such tokens can be detected automatically from the embedding matrix, because under-trained rows keep their initialization norm (or shrink under weight decay), and that the norm indicator correlates with training frequency; \citet{li2024glitch} categorise the phenomenon. We use the same signal in the opposite direction: instead of searching a vocabulary for under-trained tokens, we take a fixed, structurally interesting set of 256 tokens and ask how much exposure each has had, using the padding rows as a reference set that needs no manual selection.

\paragraph{Embedding norm and frequency.} The relationship between token frequency and embedding geometry has been studied as representation degeneration \citep{gao2019degeneration} and as the rare-token gradient problem \citep{yu2022raretokens}: frequent tokens receive more gradient and move further from initialization, while rare tokens are pushed in a common direction. We observe the monotone part of this relationship directly and also its fold at high frequency.

\paragraph{Tokenizer fairness across languages.} \citet{petrov2023unfairness} and \citet{ahia2023cost} documented that the same text costs several times more tokens in some languages. Our per-language fallback table is the extreme end of that spectrum: for scripts a vocabulary does not cover, the model is already a byte-level model.

\section{Setup}
\subsection{Tokenizers and byte-token layout}
\label{sec:layout}
We study three tokenizers: Qwen3.8 (248{,}320 rows including padding; 248{,}077 used), MiniMax-M3.1-Flash (200{,}064 rows; 200{,}061 used) and MiniCPM5-2B (130{,}560 rows, all assigned, of which 478 are \texttt{<unused\_token\_*>} reserved symbols). All three are GPT-2-style byte-level BPE with the same byte-to-unicode mapping, so each has exactly 256 single-character vocabulary entries that decode to one byte. Their ids differ: Qwen puts them at ids 0--255 and MiniCPM5 at ids 22--277, in both cases in GPT-2's byte-to-unicode order rather than by byte value; MiniMax puts them at ids 0--255 with id equal to the byte value. All scripts map ids to byte values explicitly. Qwen and MiniMax split numbers (Qwen one digit per token, MiniMax in groups of up to three digits); none of the three uses SentencePiece-style \texttt{<0xNN>} fallback.

\subsection{Corpora (route one)}
\label{sec:corpora}
All corpora are public open datasets; no private or user data were used. Each domain contributes a sample read by streaming Parquet row groups from the Hugging Face Hub (30~MB for the eight main domains, 10~MB for each additional language):
\begin{itemize}
\item \textbf{Wikipedia}: Chinese (a filtered 2023-07-20 dump) and English (10~MB sample).
\item \textbf{Web}: English from FineWeb \citep{penedo2024fineweb} (sample-10BT); Chinese and 31 other languages from the per-language test shards of FineWeb-2 \citep{penedo2025fineweb2}: de, fr, es, pt, it, pl, tr, id, vi, ru, uk, el, he, ar, fa, ur, hi, bn, ta, te, kn, ml, si, th, lo, km, my, ja, ko, ka, hy, am.
\item \textbf{Dialogue}: WildChat-1M \citep{zhao2024wildchat} (real multilingual user--model logs; user and assistant turns joined as text) and OASST2 \citep{kopf2023oasst} (human-written conversation trees).
\item \textbf{Code}: \texttt{codeparrot/github-code-clean}. \textbf{Mathematics}: FineMath-4+ \citep{allal2025smollm2}.
\end{itemize}
Each sample is tokenized without special tokens. For every token id we count occurrences; for every high-byte token we use the tokenizer's character offsets to recover the Unicode character that was split and classify it (CJK basic block, CJK extensions, Hangul, emoji, private-use area, control characters, individual scripts).

\subsection{Mixtures and totals}
\label{sec:mix}
Pretraining mixtures are not disclosed, so we form six scenarios over six buckets (English, Chinese, code, mathematics, dialogue, other languages) and report the spread (Table~\ref{tab:scen}). The ``other'' bucket is the equal-weight average of ten major languages; the \texttt{long\_tail} scenario additionally gives 10\% to the equal-weight average of the 22 further languages. Within a bucket, web text is weighted 0.8 and Wikipedia 0.2.

Totals: Qwen3.8 has not disclosed its pretraining size; the model card gives no token count or technical report \citep{qwen2026qwen38}. The Qwen3 report \citep{yang2025qwen3} states 36T tokens over 119 languages for the earlier generation, which used a smaller (151k) vocabulary. We therefore report counts \emph{per $10^{12}$ pretraining tokens} as the primary quantity and use 36T only as a scale, marked as such. MiniMax-M3.1-Flash \citep{minimax2026m31flash} and MiniCPM5-2B \citep{openbmb2026minicpm5} disclose no total either and are reported per $10^{12}$ tokens only.

\subsection{Weights (route two)}
\label{sec:weights}
We take the official bf16 checkpoints and read only the \texttt{embed\_tokens} and \texttt{lm\_head} tensors (for MiniMax, via HTTP range requests into the safetensors shards, 4.9~GB, without downloading the model). For every row we compute the $\ell_2$ norm and the norm after subtracting the column mean. Four models: Qwen3.8-27B (hidden 5120), Qwen3.5-9B (same vocabulary, hidden 4096 \citep{qwen2026qwen35_9b}), MiniCPM5-2B (hidden 2048, untied), MiniMax-M3.1-Flash (hidden 6144; embedding and output matrices are excluded from its MXFP8 quantization and stored in bf16).

\textbf{Zero-exposure baseline.} Qwen's vocabulary is padded from 248{,}077 to 248{,}320 rows; the 243 padding rows exist in the matrix but can never be selected, so they are the model's untouched initialization (possibly shrunk by weight decay). MiniMax has 3 such rows; MiniCPM5 has none, but its 478 \texttt{<unused\_token\_*>} rows play the same role. We call the median norm of these rows the baseline and its 99th percentile the baseline ceiling.

\textbf{Calibration curve.} For ordinary multi-byte tokens we know the mixture frequency $p$ from route one (base scenario). We bin $\log_{10}p$ in 0.5-dex steps and take the median norm per bin. \textbf{Equivalent frequency.} For a byte token with norm $n$, we take all ordinary tokens whose norm is within $\pm0.01$ of $n$ and report the median of their $\log_{10}p$ (``neighbourhood inversion''); this is robust to a curve that is not strictly monotone, but it is only meaningful on the monotone segment. We also report Spearman rank correlations between norm and $\log_{10}p$.

\section{Route one: corpus statistics}
\label{sec:route1}

\subsection{Single-byte tokens are mostly ASCII}
Table~\ref{tab:langs} lists bytes per token, the share of single-byte tokens and the share of high bytes for every domain and all three tokenizers. With the Qwen3.8 tokenizer, single-byte tokens are 18\% of English web text, 13\% of Chinese web text, 31\% of code and 36\% of mathematics. Digits alone are 4.7\% (English web) to 14.9\% (mathematics) of all tokens because Qwen splits every number into single digits, and ASCII punctuation adds 7--11\%. These 128 ASCII byte tokens are among the most frequent tokens in the vocabulary: the newline byte 0x0A alone would be seen $\sim10^{12}$ times at the 36T scale. Across the six mixtures the single-byte share is 18.5--21.9\%, i.e.\ about $2\times10^{11}$ occurrences per $10^{12}$ pretraining tokens.

\subsection{High bytes: 0.003\% to 45\% depending on the script}
\label{sec:langs}
High bytes appear only when a character is not in the vocabulary. For the Qwen3.8 tokenizer the rate is 0.003\% for Russian, 0.005--0.015\% for French, Italian, Turkish, German, Portuguese and English web text, 0.44\% for Chinese web text and 0.59\% for Chinese Wikipedia, 0.63\% for mathematics, 1.3\% for Korean, 1.6\% for Bengali, 1.9\% for Persian, 2.0\% for Kannada, 4.5\% for Sinhala, 6.4\% for Burmese, 9.8\% for Armenian, 12\% for Khmer, 16\% for Lao and 45\% for Amharic (Figure~\ref{fig:hb_lang}). The triggers differ by domain:
\begin{itemize}
\item \textbf{Chinese}: 82--86\% of fallbacks are rare characters of the CJK basic block that have no token (e.g.\ 讷, 厝, 冢, 腓, 谥, 祯); CJK extension characters are second.
\item \textbf{Korean}: 71\% are Hangul syllables not in the vocabulary (e.g.\ U+D53C, U+D0C0, U+D50C); Korean has the highest fallback rate of the major languages.
\item \textbf{Mathematics}: private-use-area code points (remnants of the Symbol font, e.g.\ U+F02B) 25\%, other symbols, mathematical operators and C1 control characters.
\item \textbf{Code}: C1 control characters U+0080--U+009F (39\%; typically Windows-1252 mis-decoding), then emoji.
\item \textbf{Dialogue}: emoji (25\% in WildChat), rare CJK, and pinyin tone letters (ǎ, ǐ, ǒ).
\end{itemize}

\begin{table}[htbp]
\centering\scriptsize
\setlength{\tabcolsep}{3.2pt}
\begin{tabular}{llrrrrrrrrr}
\toprule
Domain & & \multicolumn{3}{c}{Qwen3.8} & \multicolumn{3}{c}{MiniMax-M3.1-Flash} & \multicolumn{3}{c}{MiniCPM5-2B} \\
 & & B/tok & 1-byte\% & high\% & B/tok & 1-byte\% & high\% & B/tok & 1-byte\% & high\% \\
\midrule
English Wikipedia & en & 4.35 & 20.2 & 0.043 & 4.71 & 13.8 & 0.023 & 4.65 & 14.4 & 0.052 \\
English web & en & 4.43 & 18.2 & 0.015 & 4.68 & 12.0 & 0.020 & 4.62 & 12.0 & 0.040 \\
Chinese Wikipedia & zh & 3.85 & 15.9 & 0.59 & 4.18 & 7.6 & 0.64 & 4.00 & 7.7 & 0.81 \\
Chinese web & zh & 4.37 & 13.0 & 0.44 & 4.51 & 6.9 & 0.79 & 4.22 & 8.1 & 2.16 \\
GitHub code & — & 3.47 & 31.4 & 0.11 & 3.77 & 18.4 & 0.13 & 3.53 & 20.9 & 0.29 \\
\midrule
FineMath & — & 3.28 & 35.5 & 0.63 & 3.63 & 25.1 & 0.63 & 3.57 & 26.0 & 0.51 \\
\midrule
WildChat-1M & — & 4.43 & 20.5 & 0.091 & 4.61 & 14.1 & 0.087 & 4.36 & 15.8 & 0.52 \\
OASST2 & — & 4.72 & 16.1 & 0.32 & 4.79 & 11.9 & 0.39 & 4.49 & 13.1 & 0.53 \\
\midrule
German & de & 4.09 & 17.9 & 0.013 & 3.78 & 12.1 & 0.015 & 3.68 & 12.5 & 0.15 \\
French & fr & 4.11 & 15.2 & 0.005 & 4.56 & 10.1 & 0.005 & 3.82 & 9.8 & 0.033 \\
Spanish & es & 4.32 & 14.1 & 0.036 & 4.73 & 9.8 & 0.025 & 3.89 & 9.8 & 0.038 \\
Portuguese & pt & 4.21 & 16.6 & 0.014 & 4.54 & 11.6 & 0.012 & 3.80 & 11.4 & 0.051 \\
Italian & it & 4.20 & 15.4 & 0.006 & 3.73 & 11.1 & 0.009 & 3.27 & 13.4 & 0.048 \\
Polish & pl & 3.49 & 16.9 & 0.035 & 2.90 & 13.1 & 0.070 & 2.37 & 17.1 & 0.87 \\
Turkish & tr & 3.66 & 14.3 & 0.007 & 2.83 & 11.3 & 0.037 & 2.43 & 15.2 & 1.14 \\
Indonesian & id & 4.68 & 15.8 & 0.015 & 4.89 & 10.6 & 0.014 & 2.92 & 12.5 & 0.24 \\
Vietnamese & vi & 4.87 & 14.7 & 0.097 & 2.43 & 23.3 & 4.36 & 2.01 & 33.6 & 13.4 \\
Russian & ru & 6.67 & 15.8 & 0.003 & 5.41 & 8.2 & 0.004 & 5.85 & 8.9 & 0.033 \\
Ukrainian & uk & 5.04 & 12.6 & 0.060 & 4.60 & 7.3 & 0.057 & 3.68 & 11.5 & 5.62 \\
Greek & el & 4.78 & 9.1 & 0.51 & 2.67 & 5.6 & 2.60 & 2.58 & 8.7 & 5.77 \\
Hebrew & he & 3.64 & 8.3 & 0.22 & 2.27 & 15.8 & 12.3 & 1.68 & 46.5 & 43.9 \\
Arabic & ar & 5.85 & 9.0 & 0.21 & 5.87 & 4.2 & 0.14 & 2.00 & 27.7 & 22.5 \\
Persian & fa & 5.29 & 8.3 & 1.89 & 3.51 & 4.9 & 1.56 & 1.92 & 26.9 & 19.4 \\
Urdu & ur & 4.29 & 3.6 & 0.59 & 2.37 & 38.0 & 33.8 & 1.69 & 46.7 & 38.9 \\
Hindi & hi & 6.10 & 6.3 & 0.12 & 4.32 & 7.7 & 4.76 & 3.28 & 10.7 & 8.41 \\
Bengali & bn & 5.62 & 5.3 & 1.56 & 3.47 & 19.7 & 17.6 & 1.67 & 47.7 & 46.0 \\
Tamil & ta & 5.81 & 6.5 & 0.62 & 3.92 & 15.3 & 12.7 & 1.60 & 49.0 & 41.9 \\
Telugu & te & 4.45 & 6.6 & 1.12 & 3.21 & 22.8 & 20.1 & 1.48 & 53.4 & 45.7 \\
Kannada & kn & 4.31 & 6.9 & 1.98 & 1.48 & 53.3 & 46.0 & 2.18 & 40.2 & 38.6 \\
Malayalam & ml & 5.01 & 5.9 & 1.36 & 1.60 & 48.8 & 42.4 & 1.30 & 70.6 & 65.4 \\
Sinhala & si & 3.52 & 8.3 & 4.46 & 1.47 & 54.2 & 45.0 & 1.02 & 99.4 & 93.0 \\
Thai & th & 10.28 & 8.6 & 0.063 & 2.87 & 16.1 & 13.4 & 6.66 & 3.9 & 0.29 \\
Lao & lo & 2.68 & 19.8 & 16.5 & 1.39 & 63.2 & 60.0 & 1.39 & 63.2 & 60.0 \\
Khmer & km & 2.96 & 16.5 & 12.2 & 1.51 & 50.8 & 50.3 & 1.05 & 96.4 & 96.0 \\
Burmese & my & 4.22 & 8.3 & 6.42 & 1.47 & 54.4 & 53.9 & 1.47 & 54.4 & 53.9 \\
Japanese & ja & 4.99 & 10.2 & 0.29 & 5.82 & 5.4 & 0.36 & 4.39 & 6.5 & 2.36 \\
Korean & ko & 4.36 & 16.5 & 1.30 & 4.29 & 10.7 & 1.95 & 3.87 & 11.5 & 3.47 \\
Georgian & ka & 4.72 & 7.6 & 0.64 & 2.19 & 48.2 & 45.1 & 1.48 & 53.3 & 51.8 \\
Armenian & hy & 2.67 & 17.9 & 9.84 & 1.18 & 90.8 & 81.5 & 1.03 & 98.8 & 90.7 \\
Amharic & am & 1.90 & 47.8 & 45.1 & 1.13 & 89.4 & 88.3 & 1.12 & 89.9 & 88.7 \\
\bottomrule
\end{tabular}

\caption{Per-domain tokenization statistics for three tokenizers on the same public samples (Qwen3.8 / MiniMax-M3.1-Flash / MiniCPM5-2B). B/tok: UTF-8 bytes per token; 1-byte\%: share of all tokens that are single-byte tokens (mostly ASCII); high\%: share that are UTF-8 fallback bytes 0x80--0xFF. Domain sizes: 30~MB for the first eight rows, 10~MB otherwise (English Wikipedia 10~MB).}
\label{tab:langs}
\end{table}

\begin{figure}[htbp]
\centering
\begin{tikzpicture}
\begin{axis}[
  width=\textwidth, height=7.2cm,
  ybar=0pt, bar width=2.2pt,
  ymode=log, ymin=0.002, ymax=200,
  log ticks with fixed point,
  ytick={0.01,0.1,1,10,100},
  ylabel={\figHbYlabel},
  symbolic x coords={wiki_en,web_en,wiki_zh,web_zh,code_github,math_finemath,chat_wildchat,chat_oasst2,web_de,web_fr,web_es,web_pt,web_it,web_pl,web_tr,web_id,web_vi,web_ru,web_uk,web_el,web_he,web_ar,web_fa,web_ur,web_hi,web_bn,web_ta,web_te,web_kn,web_ml,web_si,web_th,web_lo,web_km,web_my,web_ja,web_ko,web_ka,web_hy,web_am},
  xtick=data, xticklabels from table={figs/hb_lang.dat}{label},
  x tick label style={rotate=60,anchor=east,font=\scriptsize},
  enlarge x limits=0.015,
  grid=major, grid style={gray!25},
  legend style={at={(0.02,0.97)},anchor=north west,font=\footnotesize,cells={anchor=west}},
  tick label style={font=\footnotesize}, label style={font=\small},
]
\addplot[fill=blue!60!black,draw=none]  table[x=domain,y=qwen38]   {figs/hb_lang.dat}; \addlegendentry{Qwen3.8}
\addplot[fill=orange!85!black,draw=none] table[x=domain,y=minimax] {figs/hb_lang.dat}; \addlegendentry{MiniMax-M3.1-Flash}
\addplot[fill=green!55!black,draw=none] table[x=domain,y=minicpm5] {figs/hb_lang.dat}; \addlegendentry{MiniCPM5-2B}
\end{axis}
\end{tikzpicture}
\caption{\figHbCaption}
\label{fig:hb_lang}
\end{figure}

The same samples under the MiniMax and MiniCPM5 tokenizers show that the fallback rate is a property of vocabulary coverage, not of the language. MiniMax covers Arabic (0.14\%) but not Hebrew (12\%), Urdu (34\%), the Indic scripts (5--46\%), Thai (13\%), Georgian (45\%) or Armenian (82\%); MiniCPM5 additionally lacks Arabic (22\%), Persian (19\%) and Hebrew (44\%), and Vietnamese (13\%) and Ukrainian (5.6\%) fall back frequently. For Lao, Khmer, Burmese, Armenian and Amharic all three tokenizers encode most characters byte by byte; for Lao and Burmese the MiniMax and MiniCPM5 numbers coincide exactly because neither vocabulary has a single token for those scripts and the pre-tokenizers agree. For such scripts these models are byte-level models already, and whatever they know about the script is stored through the high-byte embeddings.

\subsection{Mixture scenarios}
Table~\ref{tab:scen} gives the single-byte and high-byte shares under the six scenarios. For Qwen3.8 the high-byte share is 0.146\% (English-heavy) to 0.268\% (Chinese-heavy) when the ``other'' bucket contains only major languages---a spread of less than $2\times$---and 0.663\% when 10\% of the mixture is long-tail scripts. Per $10^{12}$ pretraining tokens that is $1.5$--$2.7\times10^{9}$ (mainstream) to $6.6\times10^{9}$ (long tail) high-byte tokens. At the 36T scale: $5.2$--$9.7\times10^{10}$ (mainstream) or $2.4\times10^{11}$ (long tail) in total. For MiniMax the shares are 2--3$\times$ higher (0.31--0.84\%, long tail 3.4\%) and for MiniCPM5 5$\times$ higher (0.64--1.7\%, long tail 5.3\%), because their vocabularies cover fewer scripts.

\begin{table}[htbp]
\centering\small
\begin{tabular}{lrrrrrr}
\toprule
Scenario & \multicolumn{2}{c}{Qwen3.8} & \multicolumn{2}{c}{MiniMax-M3.1-Flash} & \multicolumn{2}{c}{MiniCPM5-2B} \\
 & 1-byte\% & high\% & 1-byte\% & high\% & 1-byte\% & high\% \\
\midrule
\texttt{base} & 19.5 & 0.207 & 12.4 & 0.511 & 13.5 & 1.091 \\
\texttt{en\_heavy} & 19.9 & 0.146 & 12.8 & 0.305 & 13.6 & 0.638 \\
\texttt{zh\_heavy} & 18.5 & 0.268 & 11.5 & 0.569 & 12.6 & 1.260 \\
\texttt{code\_heavy} & 21.9 & 0.189 & 13.6 & 0.412 & 14.9 & 0.875 \\
\texttt{multilingual} & 18.5 & 0.207 & 12.3 & 0.838 & 13.6 & 1.735 \\
\texttt{long\_tail} & 18.5 & 0.663 & 14.6 & 3.376 & 16.9 & 5.263 \\
\bottomrule
\end{tabular}

\caption{Shares of single-byte tokens and of high bytes under six mixture scenarios (buckets en/zh/code/math/chat/other in \%: base 40/25/15/5/5/10; en\_heavy 60/15/12/5/3/5; zh\_heavy 30/40/12/5/5/8; code\_heavy 35/20/30/5/3/7; multilingual 35/20/12/5/3/25; long\_tail 35/20/12/5/3/15 plus 10 for 22 long-tail languages). Multiply by $10^{10}$ for counts per $10^{12}$ pretraining tokens.}
\label{tab:scen}
\end{table}

\subsection{Per-byte distribution}
\label{sec:perbyte}
Exposure is very uneven across the 128 high bytes. Under the base scenario and the Qwen3.8 tokenizer, the continuation bytes 0x80--0xBF receive 88\% of all high-byte occurrences ($6.6\times10^{10}$ at the 36T scale), two-byte lead bytes 0xC2--0xDF 7.6\%, three-byte lead bytes 0xE0--0xEF 4.3\% and four-byte lead bytes 0xF0--0xF4 0.2\%. Grouped by count at the 36T scale: 27 bytes are seen $10^{9}$--$10^{10}$ times, 55 bytes $10^{8}$--$10^{9}$, 19 bytes $10^{7}$--$10^{8}$, 11 bytes fewer than $10^{7}$ (rare two-byte leads 0xD1--0xDF, 0xE0, 0xF4), and 16 bytes never occur in the 540~MB sample: the 13 illegal bytes plus the four-byte leads 0xF1--0xF3, which only occur for code points above U+40000 (CJK Extension B and beyond). The median legal high byte is seen $6.8\times10^{8}$ times at the 36T scale, i.e.\ $1.9\times10^{7}$ per $10^{12}$ tokens. For MiniMax and MiniCPM5 the corresponding medians per $10^{12}$ tokens are $2.8\times10^{7}$ and $7.7\times10^{7}$; their never-seen sets are the 13 illegal bytes plus 0xE6, 0xE7, 0xF1--0xF3 (MiniMax) and 0xE7, 0xF1--0xF3 (MiniCPM5), i.e.\ lead bytes whose code-point ranges those vocabularies cover completely.

\section{Route two: embedding norms}
\label{sec:route2}

\subsection{Baseline, curve, and the thirteen untouched bytes}
Table~\ref{tab:models} summarises the four models and Figure~\ref{fig:curves} shows the calibration curves with the 256 byte tokens overlaid.

\begin{table}[htbp]
\centering\scriptsize
\setlength{\tabcolsep}{3.5pt}
\begin{tabular}{lrrrrlrrrr}
\toprule
Model & rows & $n_0$ & $\|e\|_0$ & $\|h\|_0$ & high bytes still at init & $\rho_e$ & $\rho_h$ & $\hat{\log p}$ & $\log p_{\rm obs}$ \\
\midrule
Qwen3.8-27B & 248,320 & 243 & 0.228 & 0.403 & 13 (12) & 0.10 & 0.55 & -5.30 & -4.69 \\
Qwen3.5-9B & 248,320 & 243 & 0.199 & 0.418 & 13 (13) & -0.65 & 0.55 & -6.41 & -4.69 \\
MiniCPM5-2B & 130,560 & 478 & 0.000 & 0.385 & 13 (0) & -0.35 & 0.64 & -5.81 & -4.03 \\
MiniMax-M3.1-Flash & 200,064 & 3 & 0.019 & 0.268 & 13 (13) & 0.79 & 0.06 & -4.44 & -4.39 \\
\bottomrule
\end{tabular}

\caption{Weight-side summary. rows: embedding rows; $n_0$: zero-exposure reference rows (padding, or unused reserved tokens for MiniCPM5); $\|e\|_0$, $\|h\|_0$: their median embedding / output-row norm; ``high bytes still at init'': number of high bytes whose embedding norm is below half the median norm of ordinary tokens (in parentheses: below the 99th percentile of the reference rows). In all four models the set is the same 13 bytes 0xC0, 0xC1, 0xF5--0xFF. $\rho_e$, $\rho_h$: Spearman correlation between corpus $\log_{10}p$ and embedding / output norm within the high bytes that occur in the corpus; $\hat{\log p}$: median neighbourhood-inverted $\log_{10}p$ of those bytes from the embedding norm; $\log p_{\rm obs}$: their median corpus $\log_{10}p$.}
\label{tab:models}
\end{table}

\begin{figure}[htbp]
\centering
\begin{tikzpicture}
\begin{groupplot}[
  group style={group size=2 by 2, horizontal sep=1.5cm, vertical sep=1.9cm, xlabels at=edge bottom, ylabels at=edge left},
  width=0.5\textwidth, height=5.0cm,
  xlabel={$\log_{10} p$}, ylabel={\figCurveYlabel},
  xmin=-9, xmax=-2,
  grid=major, grid style={gray!25},
  tick label style={font=\footnotesize}, label style={font=\small},
  title style={font=\small},
  legend style={font=\scriptsize,cells={anchor=west}},
]
\nextgroupplot[title={Qwen3.8-27B}, ymin=0, ymax=1.4, legend style={at={(0.03,0.04)},anchor=south west}]
\addplot[blue!70!black,thick,mark=*,mark size=1.2pt] table[x=logp,y=embed] {figs/curve_qwen38_27b.dat}; \addlegendentry{\figCurveLegA}
\addplot[only marks,mark=o,red!80!black,mark size=1.6pt] table[x=logp,y=embed,restrict expr to domain={\thisrow{byte}}{128:255}] {figs/bytes_qwen38_27b.dat}; \addlegendentry{\figCurveLegB}
\addplot[only marks,mark=x,gray!70!black,mark size=1.6pt] table[x=logp,y=embed,restrict expr to domain={\thisrow{byte}}{0:127}] {figs/bytes_qwen38_27b.dat}; \addlegendentry{\figCurveLegC}
\addplot[black,dashed,thin] coordinates {(-9,0.2277) (-2,0.2277)}; \addlegendentry{\figCurveLegD}
\nextgroupplot[title={Qwen3.5-9B}, ymin=0, ymax=1.2]
\addplot[blue!70!black,thick,mark=*,mark size=1.2pt] table[x=logp,y=embed] {figs/curve_qwen35_9b.dat};
\addplot[only marks,mark=o,red!80!black,mark size=1.6pt] table[x=logp,y=embed,restrict expr to domain={\thisrow{byte}}{128:255}] {figs/bytes_qwen35_9b.dat};
\addplot[only marks,mark=x,gray!70!black,mark size=1.6pt] table[x=logp,y=embed,restrict expr to domain={\thisrow{byte}}{0:127}] {figs/bytes_qwen35_9b.dat};
\addplot[black,dashed,thin] coordinates {(-9,0.1992) (-2,0.1992)};
\nextgroupplot[title={MiniCPM5-2B}, ymin=0, ymax=1.0]
\addplot[blue!70!black,thick,mark=*,mark size=1.2pt] table[x=logp,y=embed] {figs/curve_minicpm5_2b.dat};
\addplot[only marks,mark=o,red!80!black,mark size=1.6pt] table[x=logp,y=embed,restrict expr to domain={\thisrow{byte}}{128:255}] {figs/bytes_minicpm5_2b.dat};
\addplot[only marks,mark=x,gray!70!black,mark size=1.6pt] table[x=logp,y=embed,restrict expr to domain={\thisrow{byte}}{0:127}] {figs/bytes_minicpm5_2b.dat};
\addplot[black,dashed,thin] coordinates {(-9,0.0003) (-2,0.0003)};
\nextgroupplot[title={MiniMax-M3.1-Flash}, ymin=0, ymax=2.0]
\addplot[blue!70!black,thick,mark=*,mark size=1.2pt] table[x=logp,y=embed] {figs/curve_minimax_m31.dat};
\addplot[only marks,mark=o,red!80!black,mark size=1.6pt] table[x=logp,y=embed,restrict expr to domain={\thisrow{byte}}{128:255}] {figs/bytes_minimax_m31.dat};
\addplot[only marks,mark=x,gray!70!black,mark size=1.6pt] table[x=logp,y=embed,restrict expr to domain={\thisrow{byte}}{0:127}] {figs/bytes_minimax_m31.dat};
\addplot[black,dashed,thin] coordinates {(-9,0.0191) (-2,0.0191)};
\end{groupplot}
\end{tikzpicture}
\caption{\figCurveCaption}
\label{fig:curves}
\end{figure}

\textbf{Baseline.} The reference rows are extremely tight: Qwen3.8-27B 0.2277 (99th percentile 0.233), Qwen3.5-9B 0.1992, MiniMax 0.0191, and MiniCPM5 exactly 0 (its unused rows were zeroed, and all 478 share one identical output row of norm 0.385). Ordinary tokens have medians of 0.93 (Qwen3.8-27B), 0.89 (Qwen3.5-9B), 0.81 (MiniCPM5) and 1.60 (MiniMax).

\textbf{Thirteen bytes.} In all four models the high bytes whose embedding norm is far below the trained population are exactly 0xC0, 0xC1 and 0xF5--0xFF---the 13 values that cannot appear in valid UTF-8. In Qwen3.8-27B twelve of them are below the baseline ceiling and 0xFE is marginally above it (0.2332 vs 0.2330); in Qwen3.5-9B and MiniMax all thirteen are at the baseline; in MiniCPM5 their norms are 0.03--0.33 against a trained population above 0.78 (and against 0 for the zeroed reserved rows, which suggests the byte rows were initialised normally and then shrank under weight decay without ever receiving a gradient). Every legal high byte, including the four-byte leads 0xF1--0xF3 that never appear in our 540~MB sample, has a norm in the trained band (Qwen3.8-27B: median 1.07 for the 112 legal bytes seen in the sample, 0.70--0.95 for 0xF1--0xF3; MiniMax: 1.52--1.58 for the five unseen-but-legal bytes 0xE6, 0xE7, 0xF1--0xF3). The weights thus see further into the tail than a 540~MB sample does, and they confirm the one set of bytes the corpus route predicts to be absent.

\textbf{Calibration curve.} For Qwen3.8-27B the median embedding norm of ordinary tokens rises monotonically from 0.78 at $\log_{10}p=-8.25$ to 1.11 at $-3.75$ (Spearman 0.68 over all ordinary tokens) and then falls to 1.05 for the most frequent tokens. The output-row norm is uncorrelated with frequency (0.03). Qwen3.5-9B peaks earlier, at 0.96 around $\log_{10}p=-5.25$, and falls to 0.79 by $-2.75$; MiniCPM5 peaks at 0.86 around $-4.75$; MiniMax rises throughout the observed range (1.57 to 1.75) with no visible fold. The fold is the frequency-dependent part of representation degeneration \citep{gao2019degeneration,yu2022raretokens}: the most frequent tokens are pulled toward the mean direction and lose norm. Its position differs between models that share a vocabulary (Qwen3.8-27B vs Qwen3.5-9B), so it is not a property of the tokenizer.

\subsection{Equivalent frequency of the legal high bytes}
Placing the legal high bytes on their own model's curve: for Qwen3.8-27B the neighbourhood inversion gives a median $\log_{10}p=-5.30$ against a corpus value of $-4.69$ (base scenario), a difference of 0.6~dex (factor 4); for MiniMax $-4.44$ against $-4.39$ (0.05~dex). For Qwen3.5-9B and MiniCPM5 the legal high bytes lie on the falling side of the fold, so inversion returns $-6.41$ and $-5.81$ against $-4.69$ and $-4.03$---under-estimates by 1.7--1.8~dex, and the within-group rank correlation between embedding norm and frequency turns negative ($-0.65$, $-0.35$). The output-row norm behaves differently: within the high bytes its rank correlation with frequency is 0.55 in both Qwen models and 0.64 in MiniCPM5 (but 0.06 in MiniMax, where the embedding is the informative side). So each model has one matrix whose norm still tracks frequency in the high-byte band, but which one it is must be checked per model.

The ASCII bytes are the most frequent tokens of all and sit beyond the fold in every model; their norms (median 1.00 in Qwen3.8-27B) say only ``fully trained''. The eight ASCII control bytes that never occur in our samples (0x04, 0x07, 0x17--0x19, 0x1B, 0x1E, 0x1F) are also in the trained band in all models except MiniCPM5 (0.51, between its baseline 0 and its trained median 0.81), presumably because binary or mis-encoded content in pretraining contains them.

\section{Discussion}
\paragraph{What the two routes agree on.} Both routes say that in Qwen3.8 the 128 ASCII byte tokens are among the best-trained tokens in the vocabulary, that each legal high byte has been seen of order $10^{8}$--$10^{9}$ times at the 36T scale ($10^{7}$--$10^{8}$ per $10^{12}$ tokens), and that the 13 illegal bytes have never been seen. This is why byteification \citep{zhang2026onebyte} did not start from random embeddings: the 256 rows it keeps had, with 13 harmless exceptions, already received between $10^{7}$ and $10^{12}$ gradient updates each. It also explains why ASCII content survived the conversion almost untouched while multi-byte scripts needed continued training: the high bytes had been seen often, but almost always as fragments of \emph{rare} characters, so the embeddings encode ``this is the second byte of an uncommon CJK character'' rather than the identity of common characters, which the model had always read as whole tokens.

\paragraph{What this means for low-resource scripts.} For Amharic, Lao, Khmer, Burmese and Armenian, Qwen3.8 is effectively a byte-level model, and for a dozen more scripts so are MiniMax-M3.1-Flash and MiniCPM5-2B. The per-byte exposure of these scripts is then exactly the share of such text in pretraining---tiny---and it is spread over the same 128 embeddings that also carry rare Chinese characters, emoji and Symbol-font remnants. The fairness literature \citep{petrov2023unfairness,ahia2023cost} measures the cost of this in tokens; the weight route shows the other cost, that a script's entire representation has to be carried by embeddings whose gradient history is dominated by other scripts.

\paragraph{Limitations.} (i) The mixture is assumed; the sensitivity is $<2\times$ for mainstream mixtures but $3\times$ if long-tail scripts reach 10\%. (ii) Qwen3.8's pretraining size is not disclosed; all absolute counts at ``36T scale'' are the Qwen3 figure used as a scale. (iii) Samples are 10--30~MB per domain; bytes rarer than $\sim10^{-7}$ are not resolved by the corpus route (the weights resolve them qualitatively). (iv) The weights reflect pretraining and post-training together; the two cannot be separated. (v) Norm inversion is valid only on the monotone segment of a per-model curve; we report it to one significant figure in $\log_{10}p$.

\section{Conclusion}
A BPE model's 256 byte tokens are not a uniform group. The 128 ASCII bytes are ordinary, extremely frequent tokens (13--36\% of all tokens). The 115 legal high bytes are genuine UTF-8 fallback, seen in Qwen3.8 at a rate of about 0.2\% of tokens---$10^{7}$--$10^{8}$ times per $10^{12}$ pretraining tokens each---and triggered mostly by rare Chinese characters, uncovered Hangul syllables, emoji, control characters and scripts the vocabulary does not cover at all. The 13 bytes that valid UTF-8 cannot contain have never been seen, and all four models we examined say so in their weights. The two estimation routes agree to within a factor of 4 where the norm--frequency curve is monotone, and the padding rows of a vocabulary are a free, exact zero-exposure reference that any such analysis can use. Scripts, per-domain statistics and the row-norm extraction tool are released with this report.

\clearpage
\bibliographystyle{plainnat}
\bibliography{references}

\clearpage
\phantomsection\label{sec:zh}
\setcounter{section}{0}
\setcounter{table}{0}
\setcounter{figure}{0}
\setcounter{footnote}{0}
\renewcommand{\thesection}{\arabic{section}}
\renewcommand{\thetable}{\arabic{table}}
\renewcommand{\thefigure}{\arabic{figure}}
\renewcommand{\figurename}{图}
\renewcommand{\tablename}{表}
\renewcommand{\abstractname}{摘要}
\renewcommand{\refname}{参考文献}
\renewcommand{\appendixname}{附录}
\begin{center}
{\LARGE\bfseries \paperTitleZh\par}
\vspace{0.8em}
{\large \paperAuthorZh\par}
\vspace{0.4em}
2026 年 11 月
\end{center}
\begin{abstract}
byte-level BPE 词表都保留 256 个单字节 token 作为回退;把这样的模型改造成纯字节模型(我们此前对 Qwen3.8-27B 做过),起点正是这 256 行 embedding。它们在预训练里究竟被训练了多少?本文对 Qwen3.8 用两条独立路线回答并互证。路线一用 Qwen3.8 tokenizer 切分 33 个语种、40 个域、共 540~MB 的公开语料(维基、FineWeb/FineWeb-2、WildChat、OASST2、GitHub 代码、FineMath),再按几套假设的预训练配比加权。单字节 token 占全部 token 的 13--36\%,但几乎全是 ASCII(数字逐位切开、标点);真正的 UTF-8 回退字节 0x80--0xFF 在主流配比下只占 0.15--0.27\%(长尾语种占 10\% 时 0.66\%),即每 $10^{12}$ 个预训练 token 里约 $2\times10^{9}$ 次,按 Qwen3 公告的 36T 规模折算,每个合法高位字节 $10^{8}$--$10^{9}$ 次。回退率跨语种相差四个数量级(俄文 0.003\%、中文 0.44\%、韩文 1.3\%、阿姆哈拉文 45\%),由词表对各文字系统的覆盖决定。路线二读四个开源模型(Qwen3.8-27B、Qwen3.5-9B、MiniCPM5-2B、MiniMax-M3.1-Flash)embedding 与输出矩阵的行范数,以从未训练的补齐行为零曝光基线、以频次已知的普通 token 为校准曲线。四个模型里停在初始化上的,正好都是合法 UTF-8 中不可能出现的 13 个字节(0xC0、0xC1、0xF5--0xFF),其余合法高位字节全部落在高频段;从范数读出的等效频次与语料估计在 Qwen3.8-27B 上相差不到 4 倍,在 MiniMax 上相差 0.05 dex。范数--频次曲线只在模型各异的拐点以下单调,所以范数反推是逐模型、到数量级的工具。全部脚本公开;未使用任何私有数据。

\end{abstract}
\noindent\small 本文为同一篇论文的中文版;英文版在前(第 1 页起)。两版实验与数字完全一致,参考文献见英文版末。
\normalsize
\renewcommand{\figHbYlabel}{高位字节 token 占全部 token 的 \%(对数)}
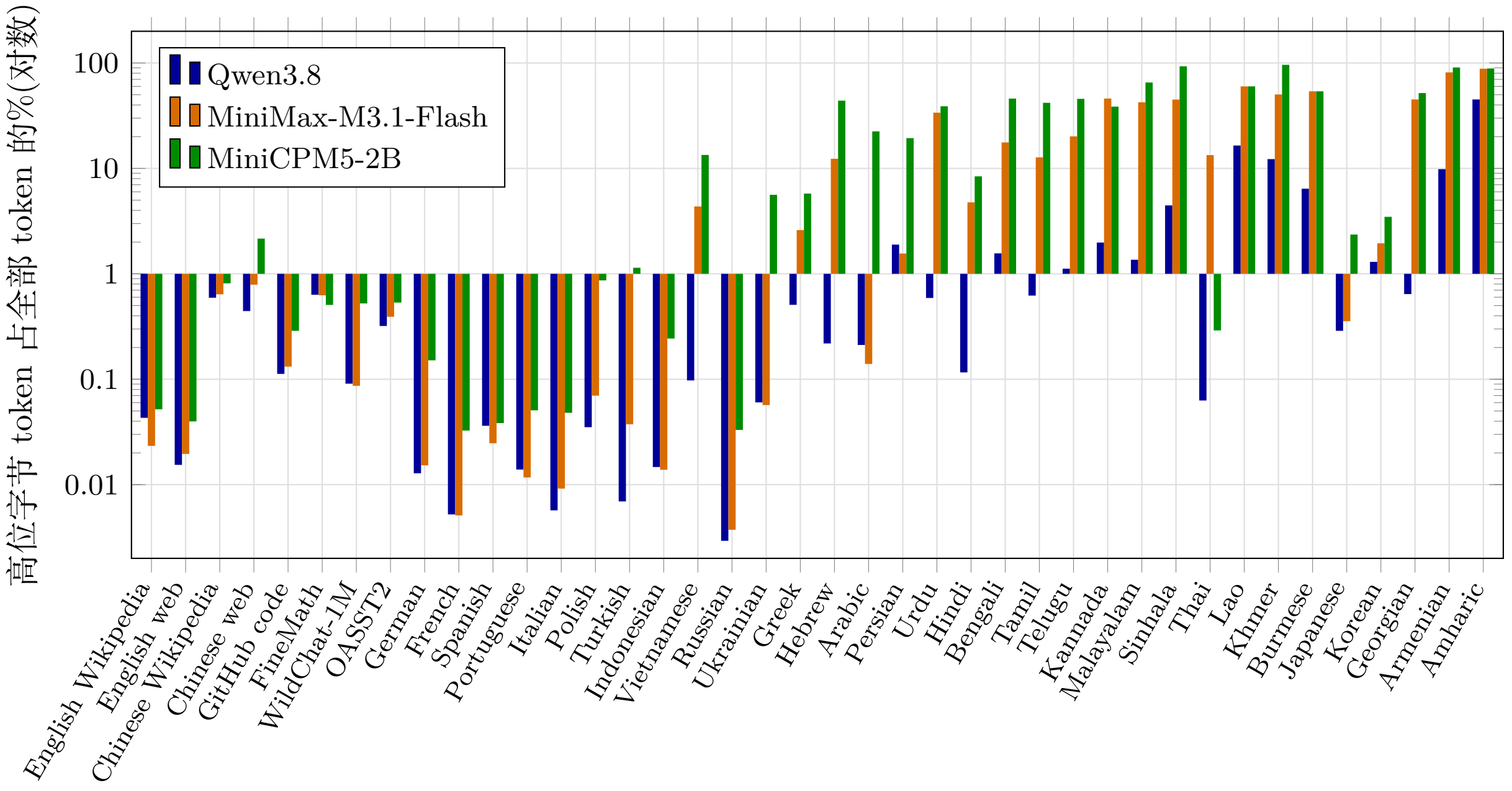
\renewcommand{\figHbCaption}{三家 tokenizer 在同一批公开语料(540~MB,33 个语种)上,UTF-8 回退字节(0x80--0xFF)占全部 token 的比例(对数坐标)。跨语种相差四个数量级;词表完全不覆盖的文字系统实际上就是按字节编码(图右侧;老挝文与缅甸文在 MiniMax 与 MiniCPM5 上数字完全相同,因为两家词表都对每个字符回退)。}
\renewcommand{\figCurveYlabel}{embedding 行范数}
\renewcommand{\figCurveLegA}{普通 token(箱中位数)}
\renewcommand{\figCurveLegB}{高位字节 0x80--0xFF}
\renewcommand{\figCurveLegC}{ASCII 字节}
\renewcommand{\figCurveLegD}{零曝光基线}
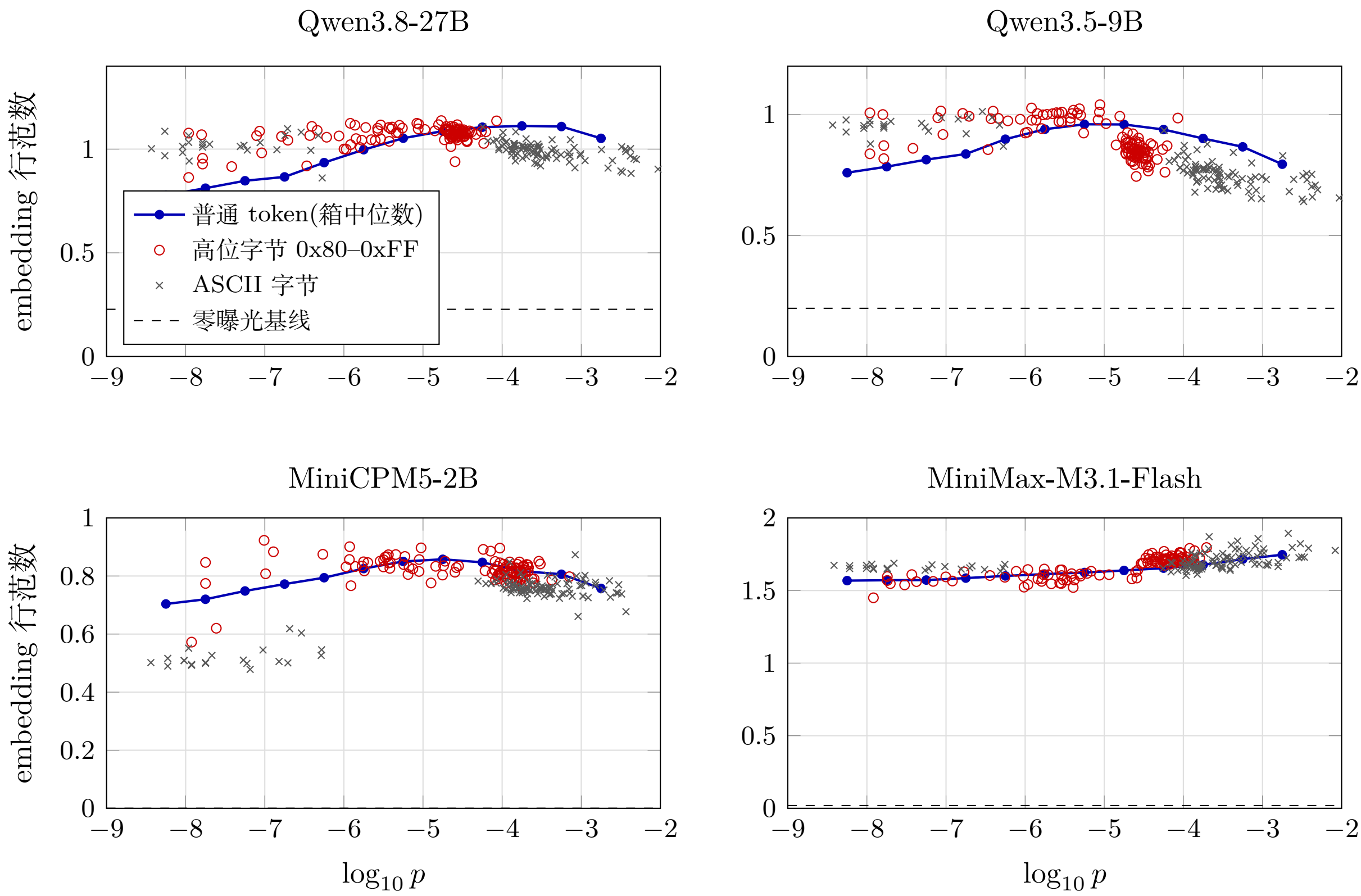
\renewcommand{\figCurveCaption}{embedding 行范数与语料频次(base 配比下 token 概率的 $\log_{10}$)。蓝线:普通多字节 token 每 0.5 dex 一箱的中位数;红圈:128 个高位字节;灰叉:128 个 ASCII 字节;虚线:从未训练的补齐行。语料样本里未出现的字节没有横坐标,只在表中列出。曲线只在模型各异的拐点以下单调(Qwen3.8-27B 约 $\log_{10}p\approx-4$,Qwen3.5-9B 与 MiniCPM5-2B 约 $-5$,MiniMax 在观测范围内没有拐点)。}

\section{引言}
每一个 GPT-2 式的 byte-level BPE 词表\citep{radford2019gpt2,sennrich2016bpe}都含有 256 个单字节 token。它们保证任何 UTF-8 字符串都能被编码:没有自己 token 的字符会被拆成字节。在此前的工作\citep{zhang2026onebyte}里,我们把 Qwen3.8-27B 改造成了纯字节模型:丢掉 embedding 与输出矩阵的其他所有行,只保留这 256 行(加特殊符号)。我们当时就注意到改造并不是"从零开始":ASCII 内容几乎没有损失,连中文也在几亿字节的训练内恢复。自然的问题是:\emph{这 256 行到底已经被训练了多少?}生产模型的预训练语料不公开,这个问题没法靠数,只能估。

本报告用两条独立路线做估计,并检查两者是否一致。
\begin{enumerate}
\item \textbf{语料路线。}用模型自己的 tokenizer 切分 33 个语种、40 个域、共 540~MB 的公开文本,统计 256 个字节 token 各出现多少次,归类触发回退的 Unicode 字符,再按几套假设的预训练配比加权。公告了预训练总量的,把占比换成次数;没公告的,按每 $10^{12}$ token 报。
\item \textbf{权重路线。}读四个开源模型 token embedding 与输出矩阵(\texttt{lm\_head})的行范数。为对齐而补齐、从未进入任何 batch 的词表行给出干净的零曝光基线;频次已由路线一给出的普通 token 给出范数--频次校准曲线;再把 256 个字节行放到曲线上。
\end{enumerate}

两组区分贯穿全文。其一是 \emph{ASCII 单字节 token}(0x00--0x7F)与\emph{高位字节}(0x80--0xFF):前者是数字、标点、空格和罕见字母的普通 token,后者只在多字节 UTF-8 字符被拆开时出现,才是真正的"字节级"信号。其二是 243 个\emph{合法}字节值与 13 个在合法 UTF-8 里根本不可能出现的字节值(0xC0、0xC1、0xF5--0xFF)。\footnote{0xC0、0xC1 会是超长的两字节首字节;0xF5--0xFF 会引出超过 U+10FFFF 的序列。128 个续字节 0x80--0xBF、首字节 0xC2--0xDF、0xE0--0xEF 与 0xF0--0xF4 都合法。}

主要结论:
\begin{itemize}
\item 每个域里单字节 token 都占全部 token 的 13--36\%,因为 Qwen 把数字逐位切开、大多数 ASCII 标点就是单字节 token;各配比下这一比例稳定在 18.5--21.9\%。这是 ASCII,不是 UTF-8 回退(第~\ref{sec:route1} 节)。
\item 真正的回退(高位字节)在主流配比下占 token 的 0.15--0.27\%,长尾文字占配比 10\% 时为 0.66\%。每 $10^{12}$ 个预训练 token 对应 $1.5$--$6.6\times10^{9}$ 个高位字节 token;按 Qwen3 公告的 36T 规模\citep{yang2025qwen3}折算,合法高位字节的中位数约被见过 $7\times10^{8}$ 次,分布从 $<10^{7}$ 到 $>10^{9}$。
\item 回退率跨语种相差四个数量级,由词表覆盖决定:俄文 0.003\%,中文 0.44\%(生僻字),韩文 1.3\%(未收录音节),缅甸文、高棉文、老挝文 6--16\%,阿姆哈拉文 45\%。词表更小的 MiniMax-M3.1-Flash 与 MiniCPM5-2B 对更多文字系统完全按字节运行(第~\ref{sec:langs} 节)。
\item 四个模型里停在初始化范数上的,正好都是那 13 个非法字节;所有合法高位字节的范数都在校准曲线的高频段(第~\ref{sec:route2} 节)。这是两条路线最硬的互证:语料里产生不出这些字节,权重确认模型也没见过它们。
\item 范数反推的等效频次与语料估计在 Qwen3.8-27B 上相差 0.6 dex、在 MiniMax 上相差 0.05 dex;但范数--频次曲线在模型各异的位置折返(Qwen3.5-9B 与 MiniCPM5-2B 折返更早),所以范数反推只是逐模型、到数量级的工具。
\end{itemize}

\section{相关工作}
\paragraph{子词词表里的字节回退。}GPT-2\citep{radford2019gpt2}引入了字节到 unicode 的映射,让 BPE\citep{sennrich2016bpe}可以在字节上运行;Qwen、MiniCPM、MiniMax 都继承了它。纯字节模型\citep{xue2022byt5,pagnoni2025blt}和把已训好模型字节化\citep{minixhofer2025bolmo,zhang2026onebyte}把字节行当成整个词表;我们问的是改造之前这些行里有什么。

\paragraph{欠训练 token 与 glitch token。}\citet{rumbelow2023solidgold}发现有些 token 会引出异常行为;\citet{land2024magikarp}证明可以从 embedding 矩阵自动检测这类 token,因为欠训练的行保留初始化范数(或在权重衰减下收缩),且范数指标与训练频次相关;\citet{li2024glitch}对现象做了分类。我们把同一信号反过来用:不是在词表里搜欠训练 token,而是取一组固定的、结构上有意义的 256 个 token,问每个被曝光了多少,并用补齐行做无需人工挑选的参照集。

\paragraph{embedding 范数与频次。}token 频次与 embedding 几何的关系被当作表示退化\citep{gao2019degeneration}和罕见 token 梯度问题\citep{yu2022raretokens}研究过:高频 token 得到更多梯度、离初始化更远,低频 token 被推向同一方向。我们直接观测到这一关系的单调段,也观测到它在高频端的折返。

\paragraph{tokenizer 的跨语种公平。}\citet{petrov2023unfairness}和\citet{ahia2023cost}记录了同一段文本在某些语言里要花几倍的 token。我们的分语种回退表是这一谱系的极端:词表不覆盖的文字系统,模型已经就是字节模型。

\section{设置}
\subsection{tokenizer 与字节 token 的布局}
\label{sec:layout}
我们研究三家 tokenizer:Qwen3.8(含补齐 248{,}320 行,已用 248{,}077)、MiniMax-M3.1-Flash(200{,}064 行,已用 200{,}061)、MiniCPM5-2B(130{,}560 行全部分配,其中 478 个是 \texttt{<unused\_token\_*>} 保留符号)。三者都是 GPT-2 式 byte-level BPE,字节到 unicode 的映射相同,所以各有恰好 256 个解码为单字节的单字符词条。id 不同:Qwen 放在 id 0--255,MiniCPM5 放在 id 22--277,两者都按 GPT-2 的字节到 unicode 顺序而非字节值排列;MiniMax 放在 id 0--255 且 id 等于字节值。所有脚本都显式地把 id 映射到字节值。Qwen 与 MiniMax 都切分数字(Qwen 一位一个 token,MiniMax 最多三位一组);三者都不用 SentencePiece 式的 \texttt{<0xNN>} 回退。

\subsection{语料(路线一)}
\label{sec:corpora}
所有语料都是公开数据集;不含任何私有或用户数据。每个域从 Hugging Face Hub 按 Parquet row group 流式抽样(8 个主要域各 30~MB,其余每个语种 10~MB):
\begin{itemize}
\item \textbf{维基}:中文(2023-07-20 过滤 dump)与英文(10~MB 样本)。
\item \textbf{网页}:英文取 FineWeb\citep{penedo2024fineweb}(sample-10BT);中文与另外 31 个语种取 FineWeb-2\citep{penedo2025fineweb2}的分语种 test 分片:de、fr、es、pt、it、pl、tr、id、vi、ru、uk、el、he、ar、fa、ur、hi、bn、ta、te、kn、ml、si、th、lo、km、my、ja、ko、ka、hy、am。
\item \textbf{对话}:WildChat-1M\citep{zhao2024wildchat}(真实多语种人机日志;用户轮与助手轮拼成文本)与 OASST2\citep{kopf2023oasst}(人写对话树)。
\item \textbf{代码}:\texttt{codeparrot/github-code-clean}。\textbf{数学}:FineMath-4+\citep{allal2025smollm2}。
\end{itemize}
每个样本不加特殊 token 切分。对每个 token id 计数;对每个高位字节 token,用 tokenizer 的字符偏移找回被拆开的 Unicode 字符并归类(CJK 基本区、CJK 扩展区、谚文、emoji、私用区、控制符、各文字系统)。

\subsection{配比与总量}
\label{sec:mix}
预训练配比不公开,所以我们在六个桶(英文、中文、代码、数学、对话、其他语种)上设六套情景,报告区间(表~\ref{tab:scen})。"其他"桶是 10 个主要语种的等权平均;\texttt{long\_tail} 情景再把 10\% 给另外 22 个语种的等权平均。桶内网页权 0.8、维基权 0.2。

总量:Qwen3.8 没有公告预训练规模,模型卡没有 token 数也没有技术报告\citep{qwen2026qwen38}。Qwen3 报告\citep{yang2025qwen3}给出上一代 119 个语种共 36T token,而那一代用的是更小的 151k 词表。因此我们以\emph{每 $10^{12}$ 预训练 token} 的次数为主量,36T 只作尺度并标明。MiniMax-M3.1-Flash\citep{minimax2026m31flash}与 MiniCPM5-2B\citep{openbmb2026minicpm5}也都没公告总量,只按每 $10^{12}$ token 报。

\subsection{权重(路线二)}
\label{sec:weights}
取官方 bf16 检查点,只读 \texttt{embed\_tokens} 与 \texttt{lm\_head} 两个张量(MiniMax 经 HTTP Range 直接读 safetensors 分片中的 4.9~GB,不下载整个模型)。每行算 $\ell_2$ 范数以及减去列均值后的范数。四个模型:Qwen3.8-27B(隐层 5120)、Qwen3.5-9B(同词表,隐层 4096\citep{qwen2026qwen35_9b})、MiniCPM5-2B(隐层 2048,不绑定)、MiniMax-M3.1-Flash(隐层 6144;embedding 与输出矩阵不在其 MXFP8 量化范围内,以 bf16 存放)。

\textbf{零曝光基线。}Qwen 的词表从 248{,}077 补齐到 248{,}320 行;这 243 行存在于矩阵里却永远不会被选中,所以是模型未动过的初始化(可能被权重衰减收缩)。MiniMax 有 3 行;MiniCPM5 没有,但它的 478 个 \texttt{<unused\_token\_*>} 行起同样作用。我们把这些行范数的中位数叫基线,第 99 百分位叫基线上限。

\textbf{校准曲线。}普通多字节 token 的配比频次 $p$ 来自路线一(base 情景)。把 $\log_{10}p$ 按 0.5 dex 分箱,取每箱范数中位数。\textbf{等效频次。}对范数为 $n$ 的字节 token,取所有范数在 $n\pm0.01$ 内的普通 token,报其 $\log_{10}p$ 的中位数("邻域反推");这对不严格单调的曲线也稳,但只在单调段有意义。同时报范数与 $\log_{10}p$ 的 Spearman 秩相关。

\section{路线一:语料统计}
\label{sec:route1}

\subsection{单字节 token 基本都是 ASCII}
表~\ref{tab:langs} 列出每个域在三家 tokenizer 下的每 token 字节数、单字节 token 占比与高位字节占比。在 Qwen3.8 tokenizer 下,单字节 token 占英文网页 18\%、中文网页 13\%、代码 31\%、数学 36\%。仅数字就占全部 token 的 4.7\%(英文网页)到 14.9\%(数学),因为 Qwen 把每个数拆成单个数字;ASCII 标点再加 7--11\%。这 128 个 ASCII 字节 token 是词表里最高频的 token 之一:换行字节 0x0A 一个,在 36T 尺度下就约被见 $10^{12}$ 次。六套配比下单字节占比 18.5--21.9\%,即每 $10^{12}$ 预训练 token 约 $2\times10^{11}$ 次。

\subsection{高位字节:随文字系统从 0.003\% 到 45\%}
\label{sec:langs}
高位字节只在字符不在词表里时出现。Qwen3.8 tokenizer 下,俄文 0.003\%,法、意、土、德、葡与英文网页 0.005--0.015\%,中文网页 0.44\%,中文维基 0.59\%,数学 0.63\%,韩文 1.3\%,孟加拉文 1.6\%,波斯文 1.9\%,卡纳达文 2.0\%,僧伽罗文 4.5\%,缅甸文 6.4\%,亚美尼亚文 9.8\%,高棉文 12\%,老挝文 16\%,阿姆哈拉文 45\%(图~\ref{fig:hb_lang})。触发成因按域不同:
\begin{itemize}
\item \textbf{中文}:82--86\% 的回退来自 CJK 基本区里没有 token 的生僻字(如 讷、厝、冢、腓、谥、祯);其次是 CJK 扩展区。
\item \textbf{韩文}:71\% 是词表未收的谚文音节(如 U+D53C、U+D0C0、U+D50C);韩文是主要语种里回退率最高的。
\item \textbf{数学}:私用区码位(Symbol 字体残留,如 U+F02B)25\%,其他符号、数学运算符与 C1 控制符。
\item \textbf{代码}:C1 控制符 U+0080--U+009F(39\%;多为 Windows-1252 误解码),其次 emoji。
\item \textbf{对话}:emoji(WildChat 中 25\%)、生僻汉字、拼音声调字母(ǎ、ǐ、ǒ)。
\end{itemize}

\begin{table}[htbp]
\centering\scriptsize
\setlength{\tabcolsep}{3.2pt}

\caption{三家 tokenizer(Qwen3.8 / MiniMax-M3.1-Flash / MiniCPM5-2B)在同一批公开样本上的分域统计。B/tok:每 token 的 UTF-8 字节数;1-byte\%:单字节 token 占全部 token 的比例(基本是 ASCII);high\%:UTF-8 回退字节 0x80--0xFF 的比例。前 8 行每域 30~MB(英文维基 10~MB),其余每语种 10~MB。}
\label{tab:langs}
\end{table}

\begin{figure}[htbp]
\centering
\begin{tikzpicture}
\begin{axis}[
  width=\textwidth, height=7.2cm,
  ybar=0pt, bar width=2.2pt,
  ymode=log, ymin=0.002, ymax=200,
  log ticks with fixed point,
  ytick={0.01,0.1,1,10,100},
  ylabel={\figHbYlabel},
  symbolic x coords={wiki_en,web_en,wiki_zh,web_zh,code_github,math_finemath,chat_wildchat,chat_oasst2,web_de,web_fr,web_es,web_pt,web_it,web_pl,web_tr,web_id,web_vi,web_ru,web_uk,web_el,web_he,web_ar,web_fa,web_ur,web_hi,web_bn,web_ta,web_te,web_kn,web_ml,web_si,web_th,web_lo,web_km,web_my,web_ja,web_ko,web_ka,web_hy,web_am},
  xtick=data, xticklabels from table={figs/hb_lang.dat}{label},
  x tick label style={rotate=60,anchor=east,font=\scriptsize},
  enlarge x limits=0.015,
  grid=major, grid style={gray!25},
  legend style={at={(0.02,0.97)},anchor=north west,font=\footnotesize,cells={anchor=west}},
  tick label style={font=\footnotesize}, label style={font=\small},
]
\addplot[fill=blue!60!black,draw=none]  table[x=domain,y=qwen38]   {figs/hb_lang.dat}; \addlegendentry{Qwen3.8}
\addplot[fill=orange!85!black,draw=none] table[x=domain,y=minimax] {figs/hb_lang.dat}; \addlegendentry{MiniMax-M3.1-Flash}
\addplot[fill=green!55!black,draw=none] table[x=domain,y=minicpm5] {figs/hb_lang.dat}; \addlegendentry{MiniCPM5-2B}
\end{axis}
\end{tikzpicture}
\caption{\figHbCaption}
\label{fig:hb_lang}
\end{figure}

同一批样本换成 MiniMax 与 MiniCPM5 的 tokenizer,说明回退率是词表覆盖的属性而不是语言的属性。MiniMax 覆盖阿拉伯文(0.14\%)但不覆盖希伯来文(12\%)、乌尔都文(34\%)、各印度系文字(5--46\%)、泰文(13\%)、格鲁吉亚文(45\%)、亚美尼亚文(82\%);MiniCPM5 还缺阿拉伯文(22\%)、波斯文(19\%)、希伯来文(44\%),越南文(13\%)和乌克兰文(5.6\%)也频繁回退。老挝文、高棉文、缅甸文、亚美尼亚文、阿姆哈拉文在三家 tokenizer 下都基本按字节编码;老挝文与缅甸文在 MiniMax 和 MiniCPM5 上数字完全一致,因为两家词表都没有一个覆盖这些文字的 token、而预切分规则相同。对这些文字,这些模型已经是字节模型,它们关于这些文字的全部知识都存放在高位字节的 embedding 里。

\subsection{配比情景}
表~\ref{tab:scen} 给出六套情景下的单字节与高位字节占比。Qwen3.8 在"其他"桶只含主要语种时,高位字节占 0.146\%(英文偏重)到 0.268\%(中文偏重),差距不到 $2\times$;长尾文字占 10\% 时为 0.663\%。每 $10^{12}$ 预训练 token 对应 $1.5$--$2.7\times10^{9}$(主流)到 $6.6\times10^{9}$(长尾)个高位字节 token。36T 尺度下合计 $5.2$--$9.7\times10^{10}$(主流)或 $2.4\times10^{11}$(长尾)。MiniMax 高 2--3 倍(0.31--0.84\%,长尾 3.4\%),MiniCPM5 高 5 倍(0.64--1.7\%,长尾 5.3\%),因为它们的词表覆盖更少文字。

\begin{table}[htbp]
\centering\small
\begin{tabular}{lrrrrrr}
\toprule
Scenario & \multicolumn{2}{c}{Qwen3.8} & \multicolumn{2}{c}{MiniMax-M3.1-Flash} & \multicolumn{2}{c}{MiniCPM5-2B} \\
 & 1-byte\% & high\% & 1-byte\% & high\% & 1-byte\% & high\% \\
\midrule
\texttt{base} & 19.5 & 0.207 & 12.4 & 0.511 & 13.5 & 1.091 \\
\texttt{en\_heavy} & 19.9 & 0.146 & 12.8 & 0.305 & 13.6 & 0.638 \\
\texttt{zh\_heavy} & 18.5 & 0.268 & 11.5 & 0.569 & 12.6 & 1.260 \\
\texttt{code\_heavy} & 21.9 & 0.189 & 13.6 & 0.412 & 14.9 & 0.875 \\
\texttt{multilingual} & 18.5 & 0.207 & 12.3 & 0.838 & 13.6 & 1.735 \\
\texttt{long\_tail} & 18.5 & 0.663 & 14.6 & 3.376 & 16.9 & 5.263 \\
\bottomrule
\end{tabular}

\caption{六套配比情景下单字节 token 与高位字节的占比(桶 en/zh/code/math/chat/other 的 \%:base 40/25/15/5/5/10;en\_heavy 60/15/12/5/3/5;zh\_heavy 30/40/12/5/5/8;code\_heavy 35/20/30/5/3/7;multilingual 35/20/12/5/3/25;long\_tail 35/20/12/5/3/15 另加 10 给 22 个长尾语种)。乘 $10^{10}$ 即每 $10^{12}$ 预训练 token 的次数。}
\label{tab:scen}
\end{table}

\subsection{逐字节分布}
\label{sec:perbyte}
128 个高位字节之间的曝光极不均匀。base 情景、Qwen3.8 tokenizer 下,续字节 0x80--0xBF 拿走全部高位字节出现次数的 88\%(36T 尺度 $6.6\times10^{10}$),两字节首 0xC2--0xDF 7.6\%,三字节首 0xE0--0xEF 4.3\%,四字节首 0xF0--0xF4 0.2\%。按 36T 尺度的次数分档:27 个字节 $10^{9}$--$10^{10}$,55 个 $10^{8}$--$10^{9}$,19 个 $10^{7}$--$10^{8}$,11 个少于 $10^{7}$(罕用的两字节首 0xD1--0xDF、0xE0、0xF4),16 个在 540~MB 样本里一次没出现:13 个非法字节加四字节首 0xF1--0xF3(只在 U+40000 以上的码位出现,即 CJK 扩展 B 及之后)。合法高位字节的中位数在 36T 尺度下被见 $6.8\times10^{8}$ 次,即每 $10^{12}$ token $1.9\times10^{7}$ 次。MiniMax 与 MiniCPM5 的对应中位数为每 $10^{12}$ token $2.8\times10^{7}$ 与 $7.7\times10^{7}$;它们未出现的集合是 13 个非法字节加 0xE6、0xE7、0xF1--0xF3(MiniMax)和 0xE7、0xF1--0xF3(MiniCPM5),即那些词表把对应码位区完全覆盖了的首字节。

\section{路线二:embedding 范数}
\label{sec:route2}

\subsection{基线、曲线与十三个没动过的字节}
表~\ref{tab:models} 汇总四个模型,图~\ref{fig:curves} 画出校准曲线并叠上 256 个字节 token。

\begin{table}[htbp]
\centering\scriptsize
\setlength{\tabcolsep}{3.5pt}
\begin{tabular}{lrrrrlrrrr}
\toprule
Model & rows & $n_0$ & $\|e\|_0$ & $\|h\|_0$ & high bytes still at init & $\rho_e$ & $\rho_h$ & $\hat{\log p}$ & $\log p_{\rm obs}$ \\
\midrule
Qwen3.8-27B & 248,320 & 243 & 0.228 & 0.403 & 13 (12) & 0.10 & 0.55 & -5.30 & -4.69 \\
Qwen3.5-9B & 248,320 & 243 & 0.199 & 0.418 & 13 (13) & -0.65 & 0.55 & -6.41 & -4.69 \\
MiniCPM5-2B & 130,560 & 478 & 0.000 & 0.385 & 13 (0) & -0.35 & 0.64 & -5.81 & -4.03 \\
MiniMax-M3.1-Flash & 200,064 & 3 & 0.019 & 0.268 & 13 (13) & 0.79 & 0.06 & -4.44 & -4.39 \\
\bottomrule
\end{tabular}

\caption{权重侧汇总。rows:embedding 行数;$n_0$:零曝光参照行(补齐行,MiniCPM5 为未用保留 token);$\|e\|_0$、$\|h\|_0$:其 embedding / 输出行范数中位数;"high bytes still at init":embedding 范数不到普通 token 中位数一半的高位字节个数(括号内:低于参照行第 99 百分位的个数)。四个模型里这一集合都是同样的 13 个字节 0xC0、0xC1、0xF5--0xFF。$\rho_e$、$\rho_h$:语料里出现过的高位字节内部,语料 $\log_{10}p$ 与 embedding / 输出范数的 Spearman 相关;$\hat{\log p}$:这些字节由 embedding 范数邻域反推的 $\log_{10}p$ 中位数;$\log p_{\rm obs}$:它们的语料 $\log_{10}p$ 中位数。}
\label{tab:models}
\end{table}

\textbf{基线。}参照行极紧:Qwen3.8-27B 0.2277(第 99 百分位 0.233),Qwen3.5-9B 0.1992,MiniMax 0.0191,MiniCPM5 恰为 0(其未用行被清零,478 行共用同一条范数 0.385 的输出行)。普通 token 的中位数分别为 0.93(Qwen3.8-27B)、0.89(Qwen3.5-9B)、0.81(MiniCPM5)、1.60(MiniMax)。

\textbf{十三个字节。}四个模型里 embedding 范数远低于已训练群体的高位字节,恰好都是 0xC0、0xC1 与 0xF5--0xFF 这 13 个合法 UTF-8 不可能出现的值。Qwen3.8-27B 中其中 12 个低于基线上限,0xFE 略高于上限(0.2332 对 0.2330);Qwen3.5-9B 与 MiniMax 中 13 个全在基线上;MiniCPM5 中它们的范数在 0.03--0.33,而已训练群体高于 0.78(被清零的保留行为 0,这说明字节行是正常初始化后在权重衰减下收缩、从未收到过梯度)。每一个合法高位字节,包括在我们 540~MB 样本里从未出现的四字节首 0xF1--0xF3,范数都在已训练区间(Qwen3.8-27B:样本里出现过的 112 个合法字节中位 1.07,0xF1--0xF3 为 0.70--0.95;MiniMax:样本里未出现但合法的 0xE6、0xE7、0xF1--0xF3 五个字节 1.52--1.58)。可见权重比 540~MB 的样本看得更远,并且恰好确认了语料路线预测为缺席的那一组字节。

\textbf{校准曲线。}Qwen3.8-27B 普通 token 的 embedding 范数中位数从 $\log_{10}p=-8.25$ 处的 0.78 单调升到 $-3.75$ 处的 1.11(全体普通 token 上 Spearman 0.68),再在最高频 token 上降到 1.05。输出行范数与频次无关(0.03)。Qwen3.5-9B 见顶更早,在 $\log_{10}p\approx-5.25$ 处 0.96,到 $-2.75$ 降至 0.79;MiniCPM5 在 $-4.75$ 处见顶 0.86;MiniMax 在整个观测范围内单调上升(1.57 到 1.75),看不到拐点。折返是表示退化\citep{gao2019degeneration,yu2022raretokens}中与频次有关的那部分:最高频的 token 被拉向均值方向而失去范数。共用词表的两个模型(Qwen3.8-27B 与 Qwen3.5-9B)拐点位置不同,所以它不是 tokenizer 的属性。

\subsection{合法高位字节的等效频次}
把合法高位字节放到各自模型的曲线上:Qwen3.8-27B 邻域反推的 $\log_{10}p$ 中位数为 $-5.30$,语料值(base 情景)为 $-4.69$,差 0.6 dex(4 倍);MiniMax 为 $-4.44$ 对 $-4.39$(0.05 dex)。Qwen3.5-9B 与 MiniCPM5 的合法高位字节落在拐点的下降侧,反推得 $-6.41$ 与 $-5.81$,对应语料值 $-4.69$ 与 $-4.03$,低估 1.7--1.8 dex,组内 embedding 范数与频次的秩相关也变成负的($-0.65$、$-0.35$)。输出行范数表现不同:高位字节内部它与频次的秩相关在两个 Qwen 模型里都是 0.55,MiniCPM5 为 0.64(但 MiniMax 只有 0.06,那里是 embedding 一侧有信息)。所以每个模型都有一侧矩阵的范数在高位字节区间仍跟着频次走,但是哪一侧必须逐模型检查。

ASCII 字节是所有 token 中最高频的,在每个模型里都在拐点之外;它们的范数(Qwen3.8-27B 中位 1.00)只能说明"训练充分"。样本里从未出现的 8 个 ASCII 控制字节(0x04、0x07、0x17--0x19、0x1B、0x1E、0x1F)在除 MiniCPM5(0.51,介于其基线 0 与已训练中位 0.81 之间)外的所有模型里也都在已训练区间,推测预训练里的二进制或错误编码内容含有它们。

\section{讨论}
\paragraph{两条路线的共识。}两条路线都说:Qwen3.8 里 128 个 ASCII 字节 token 是词表里训练得最充分的 token 之一;每个合法高位字节在 36T 尺度下被见过 $10^{8}$--$10^{9}$ 次(每 $10^{12}$ token $10^{7}$--$10^{8}$ 次);13 个非法字节从未被见过。这就是字节化\citep{zhang2026onebyte}没有从随机 embedding 起步的原因:它保留的 256 行,除 13 个无害的例外,每行已经收到过 $10^{7}$ 到 $10^{12}$ 次梯度更新。它也解释了为什么 ASCII 内容几乎原样挺过了改造,而多字节文字需要续训:高位字节虽被见过很多次,但几乎总是作为\emph{生僻}字符的碎片,所以 embedding 编码的是"这是某个不常见 CJK 字的第二个字节",而不是常用字的身份——常用字模型一向是整 token 读的。

\paragraph{对低资源文字的含义。}对阿姆哈拉文、老挝文、高棉文、缅甸文、亚美尼亚文,Qwen3.8 实际上是字节模型;MiniMax-M3.1-Flash 与 MiniCPM5-2B 对另外十几种文字也是。这些文字的逐字节曝光就是它们在预训练里的份额——极小——而且分摊在同样承载生僻汉字、emoji 与 Symbol 字体残留的那 128 个 embedding 上。公平性文献\citep{petrov2023unfairness,ahia2023cost}用 token 数衡量这一代价;权重路线显示了另一种代价:一种文字的全部表示要靠一组梯度历史被其他文字主导的 embedding 来承载。

\paragraph{局限。}(i)配比是假设;主流配比下敏感性 $<2\times$,长尾文字达到 10\% 时 $3\times$。(ii)Qwen3.8 的预训练规模未公告;所有"36T 尺度"的绝对次数都是借 Qwen3 的数字做尺度。(iii)每域样本 10--30~MB;稀于 $\sim10^{-7}$ 的字节语料路线分辨不出(权重能定性分辨)。(iv)权重反映预训练与后训练之和,两者不可分。(v)范数反推只在逐模型曲线的单调段有效;我们只报 $\log_{10}p$ 的一位有效数字。

\section{结论}
BPE 模型的 256 个字节 token 不是一个均匀的群体。128 个 ASCII 字节是普通的、极高频的 token(占全部 token 的 13--36\%)。115 个合法高位字节是真正的 UTF-8 回退,在 Qwen3.8 里约占 token 的 0.2\%——每 $10^{12}$ 预训练 token 每个字节 $10^{7}$--$10^{8}$ 次——主要由生僻汉字、未收录的谚文音节、emoji、控制符以及词表完全不覆盖的文字触发。合法 UTF-8 不可能含有的 13 个字节从未被见过,我们检查的四个模型都在权重里这样说。两条估计路线在范数--频次曲线单调的地方相差不到 4 倍;而词表的补齐行是任何此类分析都可以使用的、免费且精确的零曝光参照。脚本、分域统计与行范数提取工具随本报告公开。

\end{document}